# Toward Governance-Aware Autonomous GIS: A Narrative Review of Ethical and Privacy Risks in LLM-Enabled GeoAI

Maya Subramanian[1], Devika Jain[2*]

1. Harvard T.H. Chan School of Public Health, Boston, MA
2. Center for Geographic Analysis (CGA), Harvard University, Cambridge, MA

*Corresponding Author

Email address: kakkar@fas.harvard.edu (Devika Jain)

## Abstract:

Geospatial artificial intelligence (GeoAI) powered by large language models (LLMs) is expanding the capacity to query, generate, and interpret spatial information through natural-language interfaces and agentic "autonomous GIS" workflows. This capability is accompanied by governance challenges that general AI ethics discussions do not fully capture, including passive location inference from mobility traces, spatially structured bias amplification driven by spatial autocorrelation and scale effects, hallucinated spatial facts, and uncertainty compounding across multimodal geospatial inputs. This narrative review identifies eight recurring issues in LLM-enabled GeoAI including data provenance and consent, spatial privacy and inference risk, algorithmic bias and spatial inequity, spatial mechanisms as structural risk (spatial autocorrelation, the modifiable areal unit problem, and scale effects), LLM-specific technical risks, explainability, policy and regulatory gaps, and public enablement and workforce development. For each issue, we characterize the underlying mechanism, ground it in a concrete illustrative example drawn from the literature, and assess the current state of technical or institutional response, which ranges from largely unaddressed to actively debated to the subject of emerging policy. Building on this issue-by-issue synthesis, we propose a governance-aware architecture for LLM-enabled autonomous GIS that maps each issue to enforceable controls and auditable artifacts across the geospatial data lifecycle, illustrated through a worked flood-response routing scenario. The review highlights a persistent evidence gap: for most issues, proposed responses remain conceptual, and field-tested evaluations of governance controls for LLM-enabled GeoAI are still limited. We close by outlining a research agenda emphasizing empirical validation, spatially specific interpretability tools, and workforce training aligned to these emerging risks.

## I. Introduction and Motivation

Advances in artificial intelligence, especially breakthroughs in large language models (LLMs), are starting a new wave of innovation in the fields of spatial data science and geospatial analysis (Li et al., 2024; Xing & Sieber, 2023). Spatial analytics,the examination of spatial data, such as geographic coordinates, patterns, and relationships, helps extract insight, detect trends, and support decision-making (Miller & Goodchild, 2015). Referred to as "Geospatial Artificial Intelligence" (GeoAI), these systems blend natural language processing (NLP) with spatial analytics, opening up new ways to extract findings from heterogeneous location-based data (Li et al., 2024; Restall, 2024). What is distinct about LLMs in this context is their ability to bridge the complex geospatial data and user-friendly communications: they can interpret natural-language queries containing spatial context, synthesize insight from diverse datasets, and produce narrative explanations tailored to specific needs, capabilities reflected in emerging systems for LLM-integrated geospatial analysis (Li et al., 2024; Janowicz, 2023; Li and Ning, 2023). This enables non-technical users to interact with geospatial analysis tools in everyday language, potentially democratizing access to spatial intelligence.

At the same time, the promise of LLM-powered GeoAI is matched by a set of risks that extend beyond traditional data privacy and security concerns (Excellence and Trust in Artificial Intelligence – European Commission, n.d.; Tucker, 2024). In geospatial settings, the fusion of large spatial datasets with generative or agentic AI can intensity threats such as nuanced privacy breaches through location inference (Ntoutsi et al., 2020), amplified algorithmic bias and spatial inequities (Gupta et al., 2023), challenges in securing meaningful consent for downstream use (Restall, 2024), and the difficulty of communicating uncertainty and decision logic in complex pipelines (Xing & Sieber, 2023).

To ensure that GeoAI technologies develop in ways that align with societal values, including privacy, equity, and accountability, there is a need for structured ethical and governance frameworks tailored to LLM-integrated geospatial systems. The research landscape in this area remains fragmented, with significant gaps in empirical evidence and uneven regulatory coverage across jurisdictions. High-profile incidents have already raised concerns about privacy, ethics, and data governance; for example, sensitive location traces from fitness applications (eg. Strava) have revealed the movements of individuals at military bases. Similar failures, such as misclassification in disaster response or inequitable allocation of services, illustrate how insufficient governance in spatial AI domains can result in privacy breaches, unjust resource distribution, and erosion of trust in AI-driven systems (Ntoutsi et al., 2020; Wang et al., 2024).

This paper is a narrative review, the goal being to synthesize representative, high-relevance literature and policy discussions to (i) clarify recurring governance risks in LLM-enabled GeoAI, (ii) translate governance principles into implementable controls that can be embedded within next generation “autonomous GIS” architectures. We focus on literature from 2020 - January 2026, as it captures a rapid emergence of LLM-enabled geospatial tooling and the governance debates surrounding it.

We organize the review around eight recurring issues in LLM-enabled GeoAI governance: (i) data provenance and consent, (ii) spatial privacy and inference risk, (iii) algorithmic bias and spatial inequity, (iv) spatial mechanisms as structural risk (spatial autocorrelation, the modifiable areal unit problem, and scale effects), (v) LLM-specific technical risks, (vi) explainability and transparency, (vii) policy and regulation, and (viii) public enablement and workforce development. For each issue, we characterize the underlying mechanism, ground it in a concrete example from the literature, and assess the current state of technical or institutional response. Figure 1 below illustrates how these issues interrelate within a layered, systemic model of governance for GeoAI.

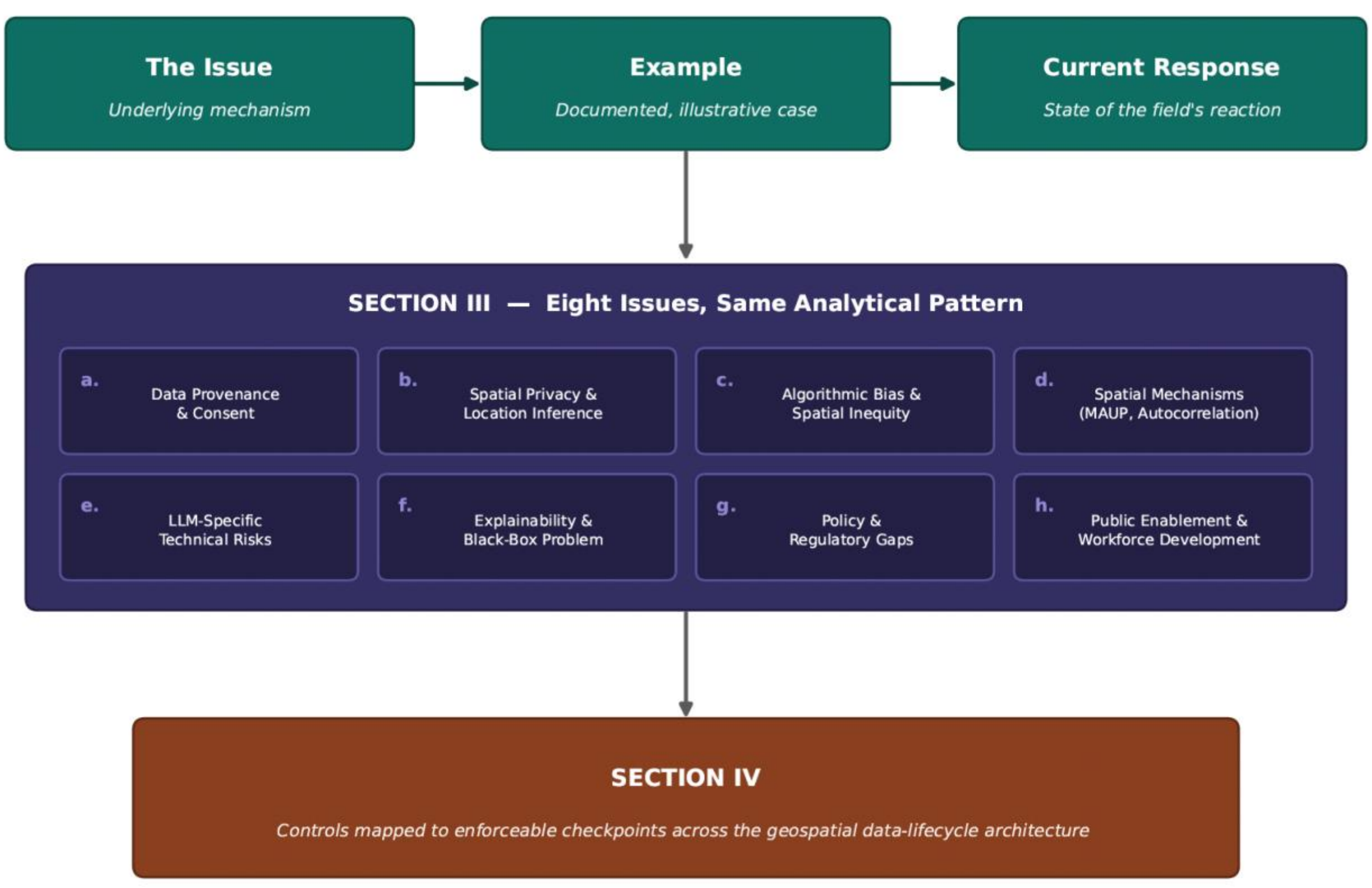

**Figure 1:** Analytical pattern applied across the eight issues in LLM-enabled GeoAI governance. This figure represents the recurring analytical pattern applied to each of the eight issues discussed in Section III: for each issue, we identify the underlying mechanism (the Issue), ground it in a concrete case (the Example), and assess the current state of technical or institutional response (Current Response). This pattern is applied uniformly across all eight issues (a through h), which then feed into Section IV, mapping issue-level findings onto enforceable controls across the geospatial data-lifecycle architecture.

This review focuses specifically on LLM-enabled GeoAI systems, including open-source tools (ChatGeoAI, UrbanGPT), and proprietary platforms. ChatGeoAI is an open-access geospatial conversational agent, helpful in supporting urban planning and spatial data queries (Wang et al., 2024). UrbanGPT, similarly, uses large language models to analyze and simulate development scenarios, helpful for policymakers and planners to evaluate proposals and social impact (Li et al., 2025). Proprietary platforms are also increasingly integrating LLMs into geospatial workflows (e.g. feature extraction, summarizing geodata, and generating spatial recommendations), which raises additional governance questions about audibility, documentation, and accountability (Hochmair et al, 2025).

Literature includes peer-reviewed articles and selected policy documents that address the integration of LLMs/AI in spatial data analysis across domains such as public health, urban planning, infrastructure, and environmental governance. Foundational literature in AI ethics and GIS is referenced where needed; however, the emphasis is on contemporary governance and risk considerations that have emerged from LLM-enabled GeoAI capabilities and autonomous GIS workflows.

## II. Methodological Approach

This narrative review synthesizes literature on ethical, privacy, and governance issues in LLM-enabled GeoAI and emerging autonomous GIS workflows. A narrative review design was selected because the topic spans heterogeneous evidence types (technical papers, conceptual governance work, and policy documents), and the field is evolving rapidly. The primary aim is integrative synthesis and operational guidance rather than exhaustive enumeration of all published studies. The review focused on literature published between January 2020 and January 2026, highlighting the recent and vast development of LLM-based GeoAI systems.

The main question guiding this review is: what are the key themes, risks, governance challenges, and policy gaps in ethics and data governance for LLM-enabled geospatial artificial intelligence (GeoAI) systems?

The purpose of this review is to map and synthesize the range and nature of literature and policy discussions addressing ethical, privacy, and governance issues in GeoAI systems that integrate large language models (LLMs), with a specific emphasis on operational controls and governance-aware architecture considerations.

Sources were identified through iterative searching and citation chaining. We conducted targeted keyword searches using Google Scholar, combining terms such as "GeoAI," "geospatial artificial intelligence," "large language models," "autonomous GIS," "spatial privacy," "location inference," "consent," "data governance," "bias," "spatial equity," "MAUP," "spatial autocorrelation," "explainable GeoAI," and "governance." We additionally used backward and forward citation chaining from seed papers central to LLM-enabled GeoAI and governance discussions (e.g., Li & Ning, 2023; Rao et al., 2023; McKenzie et al., 2023; Xing & Sieber, 2023; Wang et al., 2024). To broaden retrieval beyond initial Google Scholar results, we conducted targeted searches in Scopus and Web of Science and, where useful, directly in publisher and society portals (e.g., journal/conference digital libraries). Finally, we incorporated a limited set of policy and institutional documents referenced in the GeoAI governance discourse (e.g., OECD, European Commission, UN-GGIM, and selected industry governance materials such as IBM), treating these as governance context rather than empirical evidence. This review draws on 52 primary sources in total, of which 13 are foundational GIScience references cited for conceptual grounding rather than as literature under direct review, and 9 are policy/institutional documents. They were identified through the process described above, spanning peer-reviewed journal articles, conference proceedings, and policy/institutional documents; a subset of foundational GIScience references (e.g., Cliff & Ord, 1981; Goodchild & Gopal, 1989) are cited for conceptual grounding rather than as literature under direct review. Because this is a narrative review, we do not claim exhaustive coverage; rather, we aim for transparent synthesis of representative, high-relevance sources across the eight governance issues.

Studies were included if they:

a) Addressed any aspect of ethics, privacy, consent, bias, explainability, policy, or public enablement within LLM-powered GeoAI or NLP-based systems
b) Appeared in peer-reviewed journals, recognized conferences, or as white paper/policy documents from reputable institutions (ex. OECD, UN-GGIM)
c) Accessible in English.

Studies were excluded if they:

a) Examined tools or models without spatial or location-based components
b) Focused on general AI ethics without specific relevance to geospatial applications
c) Were not peer-reviewed or not published by reputable institutions, unless considered critical (e.g. an influential policy report from the OECD)

We organized findings into eight recurring issues in LLM-enabled GeoAI governance (Section III), each characterized by its underlying mechanism, a concrete illustrative example, and the current state of technical or institutional response. These eight issues were derived inductively: during full-text review of the literature, recurring risk mechanisms were grouped into thematic categories, which were then iteratively consolidated and refined into the eight issues presented here. Spatially specific GIS mechanisms (e.g. spatial dependence/autocorrelation, MAUP, scale effects, uncertainty propagation) constitute one such issue in their own right (Section III.d), while also surfacing as cross-cutting considerations that alter governance requirements such as auditing, validation, transparency, within the other seven.

This review is subject to several limitations in both methodology and the state of current literature. First, our inclusion criteria restricted sources to English-language, peer-reviewed articles, prominent policy documents, and select conference proceedings, which may bias findings toward predominantly English-speaking or higher-income regions. Second, because the field is rapidly evolving, recent developments, pilot projects, or ongoing deployments often appear only in grey literature or preprints and were not systematically captured; major reports from organizations such as the OECD and UN-GGIM were included where relevant, but they are not substitutes for rigorously peer-reviewed studies. Third, because many governance proposals for LLM-enabled GeoAI remain conceptual, evidence for real-world effectiveness of proposed controls is still limited; we therefore distinguish, issue by issue, between empirically supported risks (e.g., location trace re-identification) and under-validated mitigations that require applied evaluation (McKenzie et al., 2023; Wei et al., 2025).

As ethical considerations in GeoAI continue to evolve alongside technological change, the works synthesized here may become outdated, underscoring the need for ongoing updates and repeated scans of the research–policy interface in this field (Restall, 2024; Romano, 2025).

**Table 1: Evidentiary status of the illustrative example used for each of the eight issues.**

| Issue | Illustrative Evidence | Evidence Type |
|---|---|---|
| a. Data Provenance & Consent | Flood-routing / outdated map overlays | Hypothetical scenario (grounded in Roberts et al., 2023) |
| b. Spatial Privacy & Location Inference | 2018 Strava heatmap incident (Sly, 2018) | Documented incident, pre-LLM analogue |
| c. Algorithmic Bias & Spatial Inequity | Predictive policing / biased arrest data (Gupta et al., 2023) | Documented finding, pre-LLM/non-LLM analogue |
| d. Spatial Mechanisms as Structural Risk | MAUP / narrative-distortion scenarios | Established mechanism, hypothetical extension |

| Issue | Illustrative Evidence | Evidence Type |
|---|---|---|
| e. LLM-Specific Technical Risks | Flood-routing case (reused from III.a) | Hypothetical scenario (grounded in Roberts et al., 2023) |
| f. Explainability & Black-Box Problem | Xing & Sieber (2023) land-use classification case study | Documented case, non-LLM analogue |
| g. Policy Issues & Regulatory Gaps | CCPA / inferred-location regulatory gray zone | Conceptual/analytical argument |
| h. Public Enablement & Workforce Development | Municipal planner / GeoAI dashboard scenario | Hypothetical scenario |

## III. Mapping the Governance Landscape of LLM-Enabled GeoAI

### a. Data Provenance and Consent

***The Issue.*** A central concern in the ethics of GeoAI systems is governance of spatial data: particularly, how data is sourced, verified, and whether consent for said data has been appropriately obtained. Wang et. al (2024) provides a systematic review of applications of GeoAI in quantitative human geography, emphasizing how the most public geospatial datasets fundamentally lack sufficient metadata on user consent, origin, and intent. As GeoAI tools, especially those powered by LLMs, generalize vast and heterogeneous datasets, the absence of standardization in data tracing becomes increasingly problematic. While LLMs can synthesize location-based inferences and spatial decisions, they lack grounding in spatial accuracy. For instance, Lee (2025) discusses how LLMs like GPT-3 have provided plausible but incorrect outputs in health contexts. If applied in a geospatial context, this could translate into issues like misleading zoning recommendations or inaccurate disaster response plans.

***Illustrative Example (hypothetical scenario, grounded in documented LLM routing failures; see Roberts et al., 2023).*** Consider a hypothetical but realistic scenario: when planning flood response logistics, a prototype GeoAI system suggests routes through non-existent roads because of outdated map overlays. This scenario is consistent with documented empirical findings: Roberts et al. (2023) found that when GPT-4V was asked to describe walking routes from annotated OpenStreetMap images, it produced an incorrect, hallucinated route in each one of approximately 25 trials, regardless of location, marker style, or map type. When data governance protocols are weak, LLMs run the risk of ingesting and subsequently perpetuating inaccurate spatial data. This misinformation can stem from outdated maps, user-generated content, or faulty geospatial sensor feeds, all of which can distort downstream predictions of GeoAI applications.

***Current Field Response.*** Restall (2024) at IBM highlights how generative AI systems, particularly those operating on multimodal data like spatial inputs, require paradigm shifts in data governance. IBM outlines several strategies, which can all be directly applied to GeoAI: (1) enhancing data lifecycle management, ensuring that geospatial data (e.g. satellite imagery or public map contributions) are traceable from acquisition to ingestion and eventual disposal; (2) tracing model development, to understand how spatial inferences are created and evolve; (3), automating data quality and controls to get rid of outdated and/or incorrect inputs before used in high-stakes application. These remain largely proposal-stage recommendations rather than deployed standards as we are not aware of field-tested implementations of all three at once in an operational GeoAI system. Table 3 (Section IV) operationalizes these strategies as specific, auditable controls at the data-ingestion layer of the proposed governance architecture.

### b. Spatial Privacy and Location Inference

***The Issue.*** GeoAI's ability to process and synthesize such large volumes of location-based data has created unprecedented risk for privacy and surveillance violations. Rao et. al (2023) highlights the idea that privacy and security are two intertwined concepts; privacy referring to people's personal information and their rights to prevent this disclosure, while security refers to *how* that information is protected. Geospatial domains, in particular, are abundant with sensitive information, such as home addresses, workplaces, points of interest, daily trajectories, and more. Importantly, even if users do not share these details, inferences can be made from GPS tracing and digital footprints from app usage. For example, users might consent to sharing their GPS logs for fitness tracking or navigation, but not necessarily for extracting their home and work locations. However, due to the nature of GPS tracing, these key points can be identified with precision (Garroussi et al., 2025). These GPS logs are part of an increasingly growing commercial ecosystem, where data brokers such as Cubiq, SafeGraph, and Unacast can legally aggregate and sell this spatial information (Reviglio (Urbano), 2022). In turn, this data can be used in commercial and governmental ways, to estimate individual's locations without their informed understanding or consent. Much of this type of commercial activity takes place under vague user consent frameworks that fail to communicate the full extent or risks of sharing this information. Users might click “agree” to location tracking, but are rarely made aware that this same data can be purchased and recombined to create profiles. The ethics of privacy for geographic information systems have been studied for years; in 2003, the article “Geoslavery” by J.E. Dobson and P.F. Fisher highlighted the possibilities of misuse of location-based services, and how they can be used to control individuals (Dobson & Fisher, 2003). Since then, technology has only become more advanced; and the risks of privacy exposure heightened exponentially. McKenzie argues that while geospatial data can be seemingly anonymized, it can characterize individuals, their habits and patterns, and can be used for re-identification (McKenzie et al., 2023).

***Illustrative Example (documented incident, pre-LLM analogue)***. The 2018 Strava heatmap incident, in which aggregated fitness-tracking data inadvertently revealed the locations and movement patterns of individuals at military bases in active conflict zones, illustrates how passive location inference can produce consequences the original data collection never anticipated,  no single user's GPS trace was intended to reveal a base's layout, but the aggregate did (Sly, 2018).

***Current Field Response.*** Two technical approaches dominate current proposals, each with a distinct trade-off for LLM-enabled GeoAI specifically. Synthetic geospatial data generating artificial datasets that preserve the structure of real spatial data without exposing identifiable traces (Romano, 2025) offers a path to data sharing and reuse without direct privacy exposure, and has proven especially useful in domains like healthcare where obtaining real data is otherwise costly or restricted. Its limitation is a fidelity problem specific to spatial data: synthetic generation must preserve topology and adjacency, not just statistical marginals, or it risks producing "fake geography" that misrepresents rare or sparse spatial patterns that matter for equity and verifying that fidelity remains an open challenge, compounded when synthetic outputs are fed into LLM systems capable of generating superficially convincing but ungrounded narratives from them. Romano (2025) suggests engaging citizens and local actors in data-checking practices as a partial mitigation, preserving privacy benefits without fully removing human oversight from the process. Federated learning and related decentralized training approaches (Rao et al., 2023) offer the alternative of keeping raw geospatial data local rather than centralizing it, reducing exposure to large-scale breach or model-weight leakage. The limitation here is also spatially specific: geospatial data is rarely independent and identically distributed across sites, since spatial autocorrelation differs by region, which complicates convergence and can wash out the local spatial structure that federated aggregation is meant to preserve.

A related trade-off concerns transparency/traceability versus privacy in logging and lineage. The governance architecture proposed in Section IV depends on retaining audit artifacts: access logs, provenance records to make privacy protections verifiable rather than aspirational. But location logs are themselves unusually sensitive, since a query log can reconstruct the same trajectories the underlying privacy controls are meant to protect; comprehensive logging can therefore become a new attack surface rather than only a safeguard. At present, the literature offers no settled resolution to this tension — most proposals (including our own architecture in Section IV) favor generalized or access-controlled logging as a compromise, but we are not aware of empirical evaluation of how well this compromise holds up against a motivated adversary in a spatial context specifically. Rao et al. (2023) note this tension in their discussion of centralized versus decentralized foundation-model training, but frame it as an open research question rather than a solved one.

### c. Algorithmic Bias and Spatial Inequity

***The Issue.*** Algorithmic bias has increasingly become one of the forefront issues of AI ethics. Though GeoAI is being increasingly integrated into fields and systems that work to minimize adverse impacts, like infrastructure planning, public health, and more, marginalized communities are subsequently placed at risk for unintended consequences. Gupta et. al (2023) discusses the example of predictive policing in which big data systems used to learn and reproduce patterns, using historical databases. Though research has shown that these databases are not sufficient and representative of all criminal offenses, they are continually used, leading to higher rates of drug arrests in areas that are historically disadvantaged. This is due to the data used being biased, reinforcing historical beliefs that may not be fully indicative of current situations. "Fairness" is the key issue at hand: a knowledge of bias, which refers to systematic and unfair discrimination based on people's given characteristics (Gupta et al., 2023). It is also essential to consider the ways in which data collection occurred; and what kinds of bias may have occurred during that process (selection bias, exclusion bias, detection bias) (Ntoutsi et al., 2020). Several studies have analyzed the impact of multimodal ML analysis systems, and how the layers of these systems make it all the more difficult to trace the source of bias (Buolamwini & Gebru, 2018). Walker et. al (2023) expands this argument by examining how AI systems built on spatial data reflect dominant social power structures, reproducing disparities across urban-rural, racial, and economic lines. AI systems operate upon their own subjectivity, especially given the way that LLMs build upon their own datasets when they are fed more information. The issue of climate change is addressed in this article; in this case, AI’s perception of climate risk ratings and vulnerable populations are imbued with its own bias, which can exacerbate the inequalities experienced by marginalized communities. From a broader perspective, it is imperative to consider whether GeoAI is neutral. Krzysztof Janowicz (2023) argues that, operating under the principles of scalability and abstraction, algorithms operate based on their past results. Though the single steps of the GeoAI algorithms may be fundamentally neutral, networks are built up through scalability that implicitly encodes biases as demonstrated in text-to-image experiments showing inherent bias tied to geographic coverage..

***Illustrative Example (documented finding, pre-LLM/non-LLM analogue)***. Predictive policing remains the clearest documented case: historically biased arrest data, when used to train spatial risk models, produces recommendations that concentrate enforcement in the same historically over-policed neighborhoods, reinforcing rather than correcting the original disparity (Gupta et al., 2023).

***Current Field Response.*** This is currently the least operationalized issue in this review. Bias-auditing tools exist for general machine learning (e.g., the fairness metrics and disparity audits developed following Buolamwini and Gebru's 2018 work on commercial gender classification), but we found no evidence in the literature of a comparably mature, spatially explicit bias-auditing methodology that accounts for how spatial autocorrelation causes bias to cluster geographically (Section III.d) rather than distribute independently which a distinction that matters because a model can pass a global fairness check while still systematically under-serving specific neighborhoods. Walker et al. (2023) and Janowicz (2023) offer structural critique in naming the problem and its roots in historical power structures and algorithmic scalability but neither proposes a testable auditing procedure. The response at this stage is best characterized as diagnostic rather than corrective: the field has established that spatial bias is structurally distinct from general algorithmic bias, but has not yet produced deployed tools to detect or mitigate it in LLM-enabled GeoAI systems specifically. Table 3 (Section IV) proposes spatially disaggregated evaluation as a candidate control, but this remains a proposal rather than a validated practice.

### d. Spatial Mechanisms as Structural Risk

***The Issue.*** Beyond general algorithmic bias, LLM-enabled GeoAI inherits a set of spatially specific risks that arise from the structure of geographic data and spatial processes. First, spatial autocorrelation, the tendency for nearby locations to exhibit similar values, means that errors and biases are rarely independent (Cliff, 1981). Thus, when training data overrepresents types of regions or neighborhoods, models learn from these patterns and reinforce clusters of misclassification or systematically biased predictions, rather than isolated errors.

Second, the modifiable areal unit problem (MAUP) implies that results can substantially change depending on how space is partitioned and aggregated (Cooper, 2025). When LLMs summarize or explain patterns based on pre-aggregated spatial units, like census tracts or grid cells, the narratives they generate can establish the results of zoning or aggregation, rather than underlying phenomena (Chen et al., 2022). Scale effects in generative spatial reasoning arise when models are trained at one spatial/temporal resolution, but queried at another (Goodchild, 2001). LLM-based explanations can conflate neighborhood level variability with larger scale trends, masking small-area disparities that matter for health services, infrastructure, and environmental justice (Resch et al., 2025). Spatial uncertainty compounding through multimodal fusion is an emerging risk. As geospatial workflows increasingly combine remote sensing, volunteered geographic information, administrative records, uncertainties can propagate and interact (Su et al., 2024). When these sources are fused to train or prompt LLM-enabled GeoAI, it can lead to compounded uncertainty (Mohamad et al., 2026). Together, these spatial mechanisms explain why bias and error in LLM-enabled GeoAI can be structured and self-reinforcing across space, underscoring a need for governance approaches grounded in core concepts from GIScience.

Spatial autocorrelation, MAUP, and scale effects are not only conceptual concerns; they imply concrete evaluation, documentation, and uncertainty handling practices that are well established in GIScience. Work on spatial data quality and accuracy emphasizes that uncertainty arises from multiple sources (measurement error, positional inaccuracy, classification error, temporal misalignment, and processing choices), and these uncertainties can propagate through spatial operations such as overlay, buffering, and network analysis (Goodchild & Gopal, 1989; Shi et al. 2002; Devillers & Jeansoulin, 2006). Error propagation frameworks further show that when uncertain layers are combined, downstream outputs can inherit and amplify uncertainty. As a result, uncertainty should be treated as a "layer" that is quantified, tracked, and communicated rather than implicitly ignored (Heuvelink, 1998). In LLM-enabled GeoAI, these GIS principles imply governance requirements such as: (i) reporting spatial and temporal resolution/recency alongside results, (ii) documenting aggregation units and scale assumptions, (iii) producing spatially explicit error/uncertainty surfaces (not only global metrics), and (iv) conducting sensitivity analysis to determine whether conclusions are stable across zoning and aggregation choices (Openshaw, 1983; Chen et al., 2022).

***Illustrative Examples (established mechanism, hypothetical extension).*** Concrete examples can clarify why these are governance, not merely statistical, issues in LLM-enabled GeoAI.

(1) MAUP and narrative distortion: if an LLM is asked "Which neighborhoods are most effective" and the underlying analysis is aggregated at census-tract level, the resulting hotspot ranking and narrative can differ materially from a block-group or grid-based aggregation; if the system presents only one aggregation without disclosure, it can inadvertently legitimize a single zoning choice as objective truth (Openshaw, 1983; Chen et al., 2022).

    (a) Governance Implication: requires a "scale/areal-unit" declaration in outputs, and for higher stakes decisions, a brief sensitivity summary across at least one alternative aggregation.

(2) Spatial dependence and clustered harm: because spatial autocorrelation structures errors geographically, model misclassification can cluster in undermapped or underrepresented areas, creating systemic under-service that may be invisible in overall accuracy metrics (Cliff & Ord, 1981).

    (a) Governance Implication: Requires spatially disaggregating auditing (error maps, local diagnostics) rather than only one global performance metric.

(3) Scale mismatch: Models trained at one resolution (e.g. regional) but queried for neighborhood interventions can generate confident but misleading local explanations. LLM summaries may further obscure small area variability that is crucial for equity-relevant decisions (Goodchild, 2001).

(a) Governance Implication: Requires disclosure of training and evaluation scales and restricts certain claims unless validated at the requested scale.

(4) Uncertainty compounding in multimodal fusion: In flood response, uncertainty in flood extent (remote sensing or hydrologic model outputs) combined with uncertainty in road closures and map recency can propagate into routing outputs; an LLM that summarizes a "safe route" without expressing uncertainty can create a high stakes governance failure.

   (a) Governance Implication: Uncertainty propagation and user-visible confidence labeling should be required artifacts in multimodal geospatial decision support (Heuvelink, 1998; Shi et al., 2002).

***Current Field Response.*** Unlike the bias-auditing gap noted in III.c, this is an issue where established GIScience practice already provides most of the diagnostic toolkit; the response gap is one of translation and enforcement, not invention. Table 2 below summarizes each mechanism, the corresponding GIScience diagnostic or evaluation practice that already exists to detect it, and the governance requirement or artifact needed to make that diagnostic enforceable in an LLM-enabled system rather than left to the discretion of individual analysts.

**Table 2: Spatial Mechanisms as Governance Risks in LLM-enabled GeoAI: GIScience Diagnostics and Required Artifacts**

| Spatial mechanism | What can go wrong in LLM-enabled GeoAI | GIScience diagnostic / evaluation practice | Governance requirement / artifact |
|---|---|---|---|

| Spatial autocorrelation | Clustered error and inequitable harm concentration | Spatially explicit error mapping; local diagnostics; check spatial structure in residuals | Spatially disaggregated audit report; clustered error maps; mitigation triggers |
|---|---|---|---|
| MAUP | “Hotspot” narratives depend on zoning/aggregation; misleading rankings | Sensitivity analysis across alternative units/zonings | Scale/areal-unit declaration; sensitivity summary attached to outputs |
| Scale mismatch | Confident neighborhood advice from regional training; hides local disparities | Evaluate at intended decision scale; cross-scale validation | Training/evaluation scale disclosure; restrict claims without local validation |
| Multimodal uncertainty compounding | Overlay/routing outputs inherit multiple uncertainties | Error propagation / uncertainty tracking (Heuvelink tradition) | Uncertainty layer metadata; confidence labeling; uncertainty disclosure in summaries |

### e. LLM-Specific Technical Risks

***The Issue.*** The risks discussed in Sections III.a–d (bias, privacy, provenance) are not unique to LLMs: they extend concerns already present in earlier, non-generative GeoAI systems. LLM-enabled GeoAI introduces an additional layer of risk that arises specifically from how large language models generate and reason, independent of the underlying spatial data quality issues discussed above. Four mechanisms are distinct enough to warrant separate treatment.

1) **Hallucination in spatial contexts:** LLMs can generate fluent, plausible-sounding spatial claims, road names, adjacency, distances, feature existence that are not grounded in the spatial data structures they were actually queried against. This differs from generic factual hallucination in that spatial hallucinations are harder for a non-expert user to catch: a confidently described route or boundary carries no visible marker of its own unreliability the way, for instance, an obviously wrong date or name might.
2) **Prompt sensitivity and framing effects.** The same underlying spatial dataset can produce materially different LLM narratives depending on how a query specifies scale, aggregation unit, or comparison boundary. This compounds the MAUP-related narrative distortion discussed in Section III.d prompt design itself becomes a source of reproducibility risk that traditional GIS analysis, where an analyst explicitly chooses and documents an areal unit, does not have to the same degree.
3) **Multimodal grounding gaps:** When LLMs reason jointly over text, imagery, and vector/raster layers, alignment errors between modalities are not resolved by fluent output. A narrative summary can be internally consistent and confident while being unsupported by, or contradicted by, the underlying raster or vector layer it was meant to describe (Su et al., 2024; Mohamad et al., 2026).

4) **Tool and agent risk:** As GeoAI systems move toward agentic workflows that call GIS functions directly, unauthorized or unintended tool invocation becomes a distinct risk category. This is not a data-quality or bias problem but a control-plane problem: the concern is whether the system calls the tools it should, with the access it should have, not whether the tools themselves return correct results.

***Illustrative Example (hypothetical scenario, grounded in documented LLM routing failures; see Roberts et al., 2023).*** The hypothetical flood-routing scenario already introduced in Section III.a, a prototype GeoAI system routing emergency logistics through non-existent roads due to outdated map overlays, is as much a hallucination case as a data-provenance case. The same failure looks different depending on where the fault lies: if the underlying map data were accurate but the LLM still generated an unsupported route, as Roberts et al. (2023) documented empirically for GPT-4V, that would be a hallucination failure rather than a data-quality one. In practice, both failure modes can be present simultaneously and are difficult to disentangle without tool-call logging that records what the model actually queried versus what it reported.

***Current Field Response.*** Of the eight issues in this review, this is the least mature. General-purpose LLM hallucination mitigation research exists at scale, but spatially specific verification or grounding methods approaches that check a generated spatial claim against the actual GIS layer it purports to describe are largely absent from the literature we reviewed. Similarly, general prompt-robustness research exists, but nothing in the reviewed literature evaluates prompt sensitivity specifically with respect to spatial scale or aggregation-unit framing. What does exist is largely architectural rather than empirical: proposals for tool-first grounding (requiring an LLM to invoke a verifiable GIS function rather than generate a spatial claim directly) and for refusal behavior when supporting data is insufficient, both reflected as candidate controls in Table 3 (Section IV), but we found no field-tested evaluation of these controls in an operational LLM-enabled GeoAI system. This is, in our view, the most significant open gap identified in this review, a spatially explicit analogue to general LLM hallucination and robustness research was not found in the literature we reviewed, and the tool/agent risk category is new enough that even the vocabulary for evaluating it is not yet standardized. .

**f. Explainability and the Black-Box Problem**

***The Issue.*** The "black box" nature of GeoAI systems refers to the complexity of these systems, especially when understanding how outputs are derived. GeoAI models are often referred to as "opaque systems", because their innermost workings are not easily understood by users. As these models continue to grow and develop, and use previous algorithms to build new ones, it becomes harder and harder to pinpoint why outcomes are the way they are - leading to issues addressed above, such as bias, appropriate governance, and more (Pierdicca & Paolanti, 2022). Wei et al., 2025 address current explainability methods, and find that there is a lack of spatial-specific interpretability tools for LLMs. "Explainable AI" is defined as a set of processes and methods that "enable transparency and trust in outcomes of complex learning algorithms" (Wei et al., 2025). This article addresses how explainable AI can be achieved through detailed explanations, serving technical stakeholders like developers and regulators, but this can be overwhelming and confusing for users or decision-makers. Marasinghe et al., 2024 also details how most urban GeoAI tools do not provide sufficient model documentation, feedback loops, or performance metrics for stakeholder review.

***Illustrative Example (documented case, non-LLM analogue).*** Xing and Sieber (2023) provide a concrete illustration of this gap in their own land-use classification case study using a geospatial deep neural network. Applying standard XAI techniques to this task, they found the explanations produced were limited in several specific ways: appropriate reference data and models for generating the explanation itself were difficult to select, gradient-based explanation methods (the basis for techniques like Integrated Gradients) offered limited insight when applied to geographic classification tasks, and the resulting explanations struggled to accommodate geographic scale and could not be adequately visualized in map form. In other words, applying a general-purpose XAI method to a spatial task did not simply require adaptation, it surfaced failures specific to the geographic nature of the data itself.

***Current Field Response.*** One such way that explainable artificial intelligence has developed is by integrating physical models directly into model designs, making them easier to interpret for users (Zhang, n.d.). Some recent techniques produced include: LIME, SHAP, Integrated Gradients, and GNN Explainer. They have each been applied to different data types, like images, texts, and graphs, which has helped interpret the behavior of these complex models. Xing & Sieber, 2023 created a comprehensive list of the most popular algorithms and approaches for XAI, and their value for spatially explicit models. However, none of these techniques were developed with spatial reasoning in mind, and Wei et al. (2025) explicitly flag the absence of spatial-specific interpretability tools as an open gap rather than a solved problem. The general-purpose techniques above can be applied to spatial models, but they do not account for spatial structure (e.g., they would not by default surface that an explanation depends on a specific aggregation unit, the way the governance requirements in Table 2/Section III.d call for). The response here is therefore best characterized as borrowed rather than native: the field is applying general XAI tooling to spatial problems rather than developing spatial-native explainability methods.

### g. Policy Issues and Regulatory Gaps in Spatial AI

***The Issue.*** While several global regulatory frameworks exist, few are able to adequately account for the unique spatial and generative components of modern AI systems. The European Commission called for a paradigm called "Responsible AI" (*Excellence and Trust in Artificial Intelligence - European Commission*, n.d.)for new models to comply with the following principles during their design and implementation: (1) human agency and oversight, (2) technical robustness and safety, (3) privacy and data governance, (4) transparency, diversity, nondiscrimination, and fairness, (5) social and environmental wellbeing, and (6) accountability (*Excellence and Trust in Artificial Intelligence - European Commission*, n.d.). Other subsidiary pacts and policies implemented include the AI pact, to encourage voluntary commitments from companies to adhere to ethical principles, establishing regulatory "sandboxes" to experiment with AI systems in controlled environments, and placing investments into AI research and development, in areas like data surveillance and cybersecurity. However, none of these mechanisms provide binding, enforceable standards for AI system governance that derive location-specific inferences (such as those using satellite imagery, GPS logs, or spatial mobility data).

In 2023, the OECD developed the "OECD AI Principles": a set of ethical guidelines to "promote responsible and trustworthy development and use of artificial intelligence" (*AI Principles Overview - OECD.AI*, n.d.). Their framework provides general risk categories, but lacks the specific provisions needed for geospatial-based findings. Some of these spatial risks include individual re-identification via geolocation, population tracking through mobility trends, or policy applications that use inferred spatial clustering (Zang & Bolot, 2011). These gaps hinder effective governance, particularly as spatial LLMs are increasingly being deployed in large-scale settings like public health, urban development, and more.

Some emerging national frameworks, such as India's Digital Personal Data Protection Act (DPDPA), have demonstrated an evolving awareness of user rights and data localization issues, but do not highlight particular implications for spatial data (Kolanu et al., 2025). A similar pattern is visible in Brazil's Lei Geral de Proteção de Dados (LGPD): although Nakayama et al. (2025) examine its application to a different technical domain (AI-driven data sharing in retinal imaging), the same broad consent and identifiability provisions apply, and nothing in the statute specifically addresses spatial inference. Both frameworks focus primarily on personal identifiers and consent models, but do not yet articulate specific standards for passive geospatial inference or synthetic data use. At the time of this review, the U.S. has not passed comprehensive federal data privacy legislation, though proposals remain under congressional consideration; in the meantime, GeoAI regulation is left largely to the states. California, for example, enacted the California Consumer Privacy Act (CCPA), later amended by the California Privacy Rights Act (CPRA), which defines "personal information" to explicitly include both geolocation data and inferences drawn from a consumer's data (Cal. Civ. Code § 1798.140(v)(1)).

***Illustrative Example (conceptual/analytical argument).*** Even where a law formally covers inferred geospatial data, as CCPA/CPRA does, enforcement against the specific actors who assemble such profiles has been a documented gap. Geolocation and inferred characteristics are both explicitly listed as "personal information" under CCPA/CPRA, and a 2022 opinion from the California Attorney General confirmed that inferences drawn from a consumer's data qualify as personal information even when generated internally rather than collected directly. In practice, however, third-party data brokers who assemble inferred locational profiles from purchased or scraped GPS traces, rather than from a direct consumer relationship, historically posed a distinct compliance and discoverability problem: consumers had no practical way to identify which of many data brokers held inferred profiles about them. California's Delete Act (S.B. 362, 2023) was enacted specifically to address this, requiring data brokers to register with the state, disclose whether they collect precise geolocation data, and honor deletion requests through a single centralized mechanism. This example illustrates that the underlying regulatory challenge for inferred geospatial data is less a categorical gap in legal coverage and more a practical enforceability problem across a fragmented data-broker ecosystem, a distinction with direct implications for how LLM-enabled GeoAI systems that synthesize inferred spatial profiles from third-party sources should be governed.

***Current Field Response.*** A recent workshop on Responsible GeoAI emphasized the need for technical and policy integration, especially for foundational models that process remote sensing data, satellite imagery, and crowd-sourced geotags (Rao et al., 2023). Such regulatory frameworks need to prioritize traceability, issues like cross-border data transfer restrictions, and model registries to monitor bias in making spatial predictions. When addressing this issue at an international level, it's important to consider how laws would treat cross-border geospatial inferences. Multilateral organizations like the UN-GGIM have issued general policy risk briefs, but are too insufficient to address LLM-specific risks (*UNSD — UN-GGIM*, n.d.). Underlying much of this policy discussion is an unresolved institutional question: whether governance of spatially inferred data is better handled centrally or in a decentralized, jurisdiction-by-jurisdiction manner. Centralized governance, a single binding standard, comparable to the EU's approach, offers consistency and easier enforcement, but geospatial data routinely crosses jurisdictional lines (a satellite image or mobility dataset does not respect national borders the way a domestic privacy law does), which a purely centralized model struggles to address without international buy-in. Decentralized governance each jurisdiction regulating spatial inference on its own terms, as the current DPDPA/LGPD/CCPA patchwork effectively does, is more achievable in the near term but produces exactly the gray zones described above, since inferred geospatial data can simply be processed or stored in whichever jurisdiction regulates it least. Neither model has been demonstrated at scale for spatial data specifically; the Responsible GeoAI workshop's call for model registries (Rao et al., 2023) is one proposal for a middle path, since a registry could operate as shared infrastructure that either governance model could draw on, but this too remains a proposal rather than an implemented practice.

### h. Public Enablement and Workforce Development

***The Issue.*** User education plays a foundational role in establishing responsible, transparent, and equitable use protocols. At the core of governance, technical and regulatory measures remain all-important, but this must be supplemented with literacy and training programs to ensure users understand the implications of using spatially-intelligent systems (Floridi et al., 2018).

Increasingly, general AI tools, but also GeoAI tools have been reaching non-expert users, in all manner of fields (urban planners, administrators, organization leaders, etc), to make map-based visualizations and data-based conclusions. However, there is a general lack of awareness and knowledge on issues of data privacy, algorithmic ideas, and spatial inference. According to the IBM Institute of Business Value (2024), it was found that 79% of public sector leaders reported a lack of user education being a major barrier to adoption of building trustworthy AI systems. Without clear resources, users can also easily misinterpret results, contribute location data unknowingly, or fail to recognize how their input might be aggregated,  a risk that compounds every issue discussed in Sections III.a–g, since none of those governance mechanisms function if the people relying on them don't understand what they're being shown. This literacy gap has a second, less-discussed dimension: it also affects the people who *build* these systems, not only those who use their outputs. GIS curricula have historically emphasized spatial analysis, cartographic reasoning, and uncertainty/error-propagation methods (Goodchild & Gopal, 1989; Shi et al., 2002), but have not yet incorporated LLM-specific risks such as hallucination, prompt sensitivity, tool/agent risk (Section III.e)  as a standard part of professional GIS training.

***Illustrative Example (hypothetical scenario).*** Consider a municipal planner using a GeoAI dashboard to identify underserved neighborhoods for a new service, without understanding that the "underserved" ranking depends on an aggregation-unit choice made upstream in the system (Section III.d's MAUP discussion). The planner is not equipped to ask the one question that would reveal the ranking's fragility whether the same neighborhoods would rank as underserved under a different areal-unit choice not because the information is hidden, but because neither general AI literacy nor traditional GIS training currently emphasizes this as a question to ask of an LLM-mediated output specifically.

***Current Field Response:*** To address this, IBM recommended an integration of “AI Literacy Models” into workplace policy training (Restall, 2024). The OECD AI Principles identify public understanding of AI systems, including spatial data systems, as foundational to building trust and supporting equitable outcomes, though we are not aware of empirical studies directly measuring this relationship in the GeoAI context (*AI Principles Overview - OECD.AI*, n.d.). A systematic review on geospatial embedding practices in LLMs proposed a framework that included localized consent forms, user dashboards, and other heightened transparency measures – equipping users with real-time visibility data into what data is influencing their decisions, and the mechanisms by which that is happening (Tucker, 2024).

Thus, GeoAI systems, particularly those that use integrated LLMs, should be accompanied by tiered enablement plans, with involvement at the government, NGO, and general public levels, supported with open-access materials that address the above issues. Without these, even with governance and legal mechanisms in place, frameworks will fall short in protecting users and ensuring equitable participation in AI decision-making. On the workforce-training side, the response is considerably less developed. We are not aware of published GIS curricula that formally incorporate LLM-specific risks alongside traditional uncertainty and error-propagation training, despite the underlying diagnostic skills recognizing scale mismatch, auditing for spatially clustered error transferring directly from one context to the other. "Spatial AI auditing" is emerging as a distinct skill set, combining GIScience diagnostic practice with the control-mapping approach proposed in Table 3 (Section IV), but it does not yet have a home in standard GIS education, nor a competency standard defining it. Professional bodies such as URISA, the GIS Certification Institute, and the American Association of Geographers are positioned to develop such standards comparable to the "spatial ethics training" principle already named in our governance framework (Figure 2) but as of this review, we found no evidence that any has done so specifically for LLM-enabled GeoAI.

## IV. Toward an Operational Governance Architecture

The eight issues discussed in Section III are, individually, mostly proposal-stage: each has some technical or institutional response, but almost none has been field-tested, and several LLM-specific technical risks (III.e) and spatially-explicit bias auditing (III.c) chief among them have barely moved past naming the problem. This section proposes how these fragmented, issue-level responses could be unified within a single governance-aware architecture for autonomous GIS, rather than left as disconnected proposals attached to separate issues.

Before doing so, it is worth situating this contribution against three adjacent bodies of work. General AI-ethics frameworks Floridi et al.'s (2018) AI4People, the OECD AI Principles (n.d.), the EU's Responsible AI principles  articulate principles such as fairness, transparency, and accountability, but are spatially agnostic: they do not specify *where in a spatial data lifecycle* a given principle should be enforced, which is precisely the gap Section III documented issue by issue. Existing GeoAI ethics literature (Gupta et al., 2023; McKenzie et al., 2023; Ntoutsi et al., 2020) identifies spatial risks in detail but does not propose an architectural mapping of controls to system layers. Autonomous GIS architecture papers (Li & Ning, 2023; Li et al., 2025) describe system layers in technical terms but do not embed governance as a first-class layer within them. Our contribution is not a new set of ethical principles, Section III drew on existing ones throughout, but the explicit link between the issue-level mechanisms documented there and an auditable architecture: a mapping from risk mechanism to enforceable control to verifiable artifact, specific to the spatial data lifecycle. This also extends to operational AI risk-management standards not discussed above, specifically the NIST AI Risk Management Framework (National Institute of Standards and Technology [NIST], 2023) and ISO/IEC 42001:2023 (International Organization for Standardization [ISO], 2023), which are more mature and directly enforceable than principle-level frameworks but do not address spatial or LLM-specific risks at all. This comparison is elaborated in full in Table 5, once the proposed architecture and its controls have been introduced.

Building on recent work on autonomous GIS (Li & Ning, 2023; Li et al., 2025), we propose a simple conceptual architecture for LLM-integrated GIS as shown in Figure 2. At the base, a data layer manages heterogeneous spatial inputs (such as vector and raster data, trajectories, and point clouds) together with basic provenance and consent metadata. Above this, a GIS processing layer provides core spatial functions, including indexing, overlay, buffering, and network analysis. On top of these capabilities, a GeoAI/LLM layer combines spatial features with textual and contextual inputs to enable natural-language querying, scenario generation, and other generative tasks. Finally, an application and interaction layer exposes these functions through dashboards, APIs, and conversational interfaces for planners, analysts, and other end users. This layered view is intentionally high-level, but it helps clarify where ethical and governance controls can be embedded within LLM-enabled GIS, rather than treated as purely external procedures. Within this architecture, governance can be expressed as checkpoints along the spatial data lifecycle. At ingestion, systems should record provenance, consent status, user restrictions, and screening for sensitive spatial content (McKenzie et al., 2023; Wang et al., 2024). During storage and integration, controls on linkage, re-identification risk, and cross-border transfer are crucial (Dobson & Fisher, 2003; Zang & Bolot, 2011). In the modeling and training phase, documenting training, synthetic data generation, and fine-tuning procedures enables later auditing of bias and privacy risk (Ntoutsi et al., 2020; Rao et al., 2023). At the stage of interference and deployment, logging, access control, and in-time review is important for high stakes applications like zoning, policing, or disaster response (Gupta et al., 2023; Walker et al., 2023). Finally, decommissioning stages should specify deletion, retention, or repurposing of data and models, aligning with governance principles laid out for GeoAI (Floridi et al., 2018;  Hochmair et al., 2025).

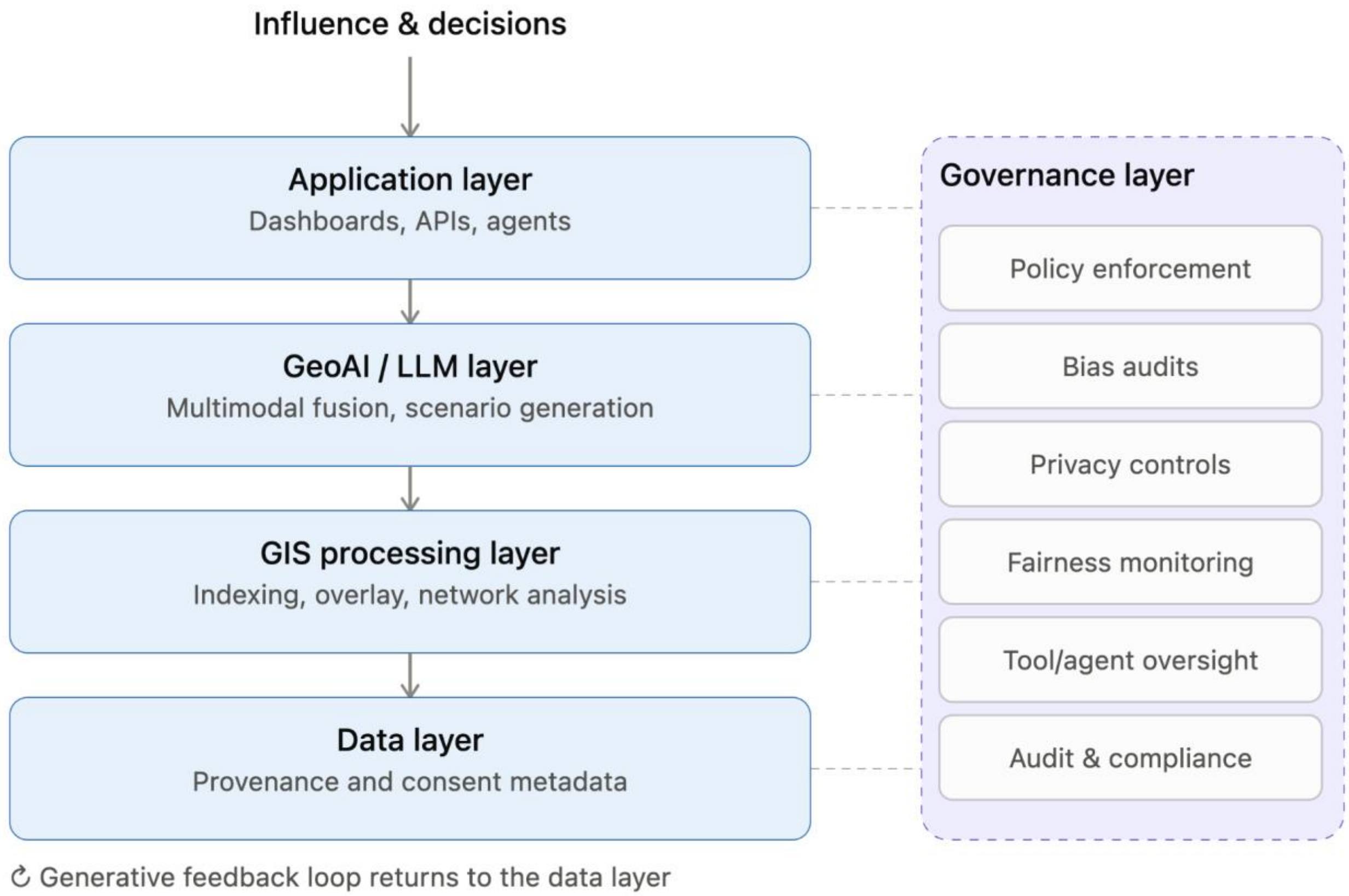

**Figure 2:** Governance aware architecture for LLM-enabled Autonomous GIS

LLM-enabled GeoAI introduces new pathways for risk propagation. To operationalize the governance checkpoints implied by the layered architecture (Figure 2), Table 3 maps common risks in LLM-enabled GeoAI to implementable controls, indicates where each control is enforced in the architecture, and lists auditable artifacts that provide evidence of implementation. Each control is additionally labeled by validation status: 'Candidate' refers to a control proposed as part of this architecture that has not been empirically validated or field-tested in an operational LLM-enabled GeoAI system; 'Established practice' refers to a control that operationalizes a technique already documented/validated in the GIScience literature (Table 2), applied here as a formal governance mechanism. This distinction is maintained consistently across Tables 3 and 4.

**Table 3: Operational Governance Controls Mapped to the LLM-enabled autonomous GIS architecture (Figure 2)**

| Risk/Failure Mode | Implementable Control | Where enforced (Figure 2 layer) | Audit Artifact | Validation Status |
| --- | --- | --- | --- | --- |
| Outdated / low-quality base maps, stale hazard layers | Enforce data recency thresholds; automated QA checks at ingestion | Data layer + GIS processing layer | Dataset registry entry; ingestion log; "recency status" flags | Candidate (proposed here; not field-tested) |

| Risk/Failure Mode | Implementable Control | Where enforced (Figure 2 layer) | Audit Artifact | Validation Status |
|---|---|---|---|---|
| Missing/unclear consent and use constraints | Machine-readable usage constraints; query-time enforcement | Data layer + Embedded governance layer | Consent receipt/record; access decision logs | Candidate (proposed here; not field-tested) |
| Passive location inference / re-identification risk | Inference-risk testing; aggregation/minimization defaults for sensitive layers | GeoAI/LLM layer + Governance layer | Privacy test report; configuration record | Candidate (proposed here; not field-tested) |
| Hallucinated spatial facts (roads, closures, boundaries) | Tool-first grounding (LLM must call routing/overlay tools); refusal behavior when insufficient data | GeoAI/LLM layer | Tool-call logs; refusal logs; output provenance record | Candidate (proposed here; not empirically validated) |

| Risk/Failure Mode | Implementable Control | Where enforced (Figure 2 layer) | Audit Artifact | Validation Status |
|---|---|---|---|---|
| Prompt sensitivity / hidden assumptions (scale, units, travel mode) | Standard prompt templates + forced parameter disclosure | App/UI layer + GeoAI/LLM layer | Prompt template ID; parameter snapshot; prompt/version log | Candidate (proposed here; not field-tested) |
| MAUP / aggregation-driven narrative distortion | Scale/areal-unit declaration + sensitivity check across at least one alternative unit | GIS processing layer + App/UI | Scale annotation stored with outputs; sensitivity summary | Established practice (operationalizes documented GIScience diagnostic; see Table 2) |
| Spatially clustered error / inequitable impact | Spatially disaggregated evaluation; threshold triggers + mitigation workflow | Governance layer + GIS processing | Spatial error maps; disparity metrics report; mitigation action log | Established practice (operationalizes documented GIScience diagnostic; see Table 2) |

| Risk/Failure Mode | Implementable Control | Where enforced (Figure 2 layer) | Audit Artifact | Validation Status |
|---|---|---|---|---|
| Multimodal uncertainty compounding | Uncertainty propagation tracking + confidence labeling in outputs | GIS processing + App/UI | Uncertainty fields; confidence labels; uncertainty disclosure text | Candidate (proposed here; not field-tested) |
| Tool misuse in agentic workflows (unauthorized API calls) | Tool allow-list; RBAC; monitored tool calls | GeoAI/LLM layer + Governance layer | Tool allow-list policy; tool-call audit logs; access audit | Candidate (proposed here; not field-tested) |
| Lack of contestability | User contestation workflow + escalation roles | App/UI layer + Governance layer | Contestation tickets; reviewer decisions; corrective action record | Candidate (proposed here; not field-tested) |

Bias or error introduced early, like under-representation of certain regions, incomplete consent metadata, or outdated base maps, can be amplified by downstream models and then naturalized (Gupta et al., 2023; Ntoutsi et al., 2020). In domains such as predictive policing, climate-risk assessment, or infrastructure planning, such compounded errors may translate into systematically unfair or unsafe recommendations (Walker et al., 2023; Marasinghe et al., 2024). Viewing LLM-enabled GIS as a chain of components (data sources, preprocessing steps, spatial models, LLM prompts, and user interfaces), helps identify where privacy leakage, bias amplification, or loss of spatial context are likely (Pierdicca & Paolanti, 2022; Rao et al., 2023). Governance-aware architectures should therefore include explicit risk assessment and mitigation hooks at each stage, including impact analysis for vulnerable populations and mechanisms for contesting harmful outputs (Janowicz, 2023; Romano, 2025).

The "hallucinated spatial facts" row above reflects a broader design trade-off worth naming explicitly: grounded, tool-first GIS generation versus free-form LLM output. Tool-first generation, requiring the LLM to invoke a verifiable GIS function (routing, overlay) rather than generate a spatial claim directly, is theoretically motivated to reduce hallucination risk. This is because outputs are computed rather than asserted, but this has not been empirically evaluated in an operational LLM-enabled GeoAI system. Further, tool-first generation constrains response flexibility and requires more upfront engineering investment. Free-form generation is faster and requires less engineering, but carries the hallucination risk illustrated by the flood-routing scenario (Section III.a) and documented empirically in LLM map-routing tasks more broadly (Roberts et al., 2023). Because spatial correctness is frequently safety-critical, a hallucinated route is not a cosmetic error the way a hallucinated fact might be in a general-purpose chatbot; we regard tool-first grounding as the appropriate default for high-stakes GeoAI applications, with free-form generation reserved for lower-stakes, exploratory use.’

**Illustrative Operational Scenario: LLM-enabled flood response routing agent**

- Context-goal: Deploying an LLM-enabled GIS assistant to support flood response during rapidly evolving event
- System's purpose is to (i) generate safe routing options for emergency vehicles/evacuees, (ii) situational summaries (e.g. impacted road segments, neighborhoods with reduced access, (iii): resource allocation recommendations.

Because these outputs can directly affect safety and equity, the system is configured as decision support with explicit governance controls, rather than an autonomous decision maker.

**System components mapped to the architecture:**

- Data layer inputs (road networks/restrictions, critical facilities, mobility signals like traffic speeds, 911 call density, etc)
- GIS processing layer functions: network cleaning/topology checks, overlay flood surfaces, passability constraints, network routing
- GeoAI/LLM layer: translating user questions into auditable GIS tool calls, summarizing results into plain language, generating structured action lists with uncertainty disclosures
- Application layer: dispatcher dashboard, conversational interface w/ controlled prompt templates
- Embedded governance services: provenance/consent enforcement, audit logging, privacy risk checks, spatial validation checks

This illustrative scenario shows how the architecture in Figure 2 can host enforceable governance controls, plus institutional roles that maintain accountability in a high stakes setting.

The interaction between spatial data structures and LLM-based generative outputs is bidirectional. Internally, LLM-integrated GIS operates over grids, graphs, indexes, and knowledge graphs. Design choices about how these structures are exposed to the LLM shape which spatial relationships can thus be learned or explained, a dynamic illustrated by LLM-integrated spatio-temporal systems more broadly (Li et al., 2024; Pierdicca & Paolanti, 2022). Externally, generative outputs, like narratives, policy briefs, or synthetic geographies, may be written back into spatial databases as new layers or annotations, reshaping the geospatial knowledge base on which subsequent models train (Hochmair et al., 2025; Romano, 2025). This feedback loop raises concerns about semantic drift, accumulation of model-generated artefacts, and erosion of the distinction between authoritative and synthetic geodata (McKenzie et al., 2023; Tucker, 2024). Governance mechanisms therefore need to regulate when and how LLM-generated content can modify spatial data stores, and under what validation or review procedures these changes are accepted (Rao et al., 2023; Floridi et al., 2018).

These considerations motivate treating governance as a core architectural layer in next-generation GIS, rather than a separate compliance afterthought. In practice, this implies encoding consent rules, usage constraints, and jurisdictional limits as machine-readable policies enforced at runtime across the architecture (AI Principles Overview—OECD.AI, n.d.; Excellence and Trust in Artificial Intelligence—European Commission, n.d.). It also calls for integrating auditing, monitoring, and bias-detection services as shared infrastructure modules, rather than add-ons to individual applications (Ntoutsi et al., 2020; Wei et al., 2025). For autonomous GIS systems that can initiate data collection, model retraining, and decision support with limited human oversight, an embedded governance layer is essential to limit risk and maintain accountability (Li & Ning, 2023; Li et al., 2025). Framing governance in architectural terms aligns GIScience with broader efforts to build "responsible" GeoAI infrastructures that are technically robust, socially legitimate, and responsive to evolving regulatory norms (Floridi et al., 2018; Xing & Sieber, 2023).

In parallel with the empirical and technical priorities addressed in this review, future work must focus on the operationalization of ethical frameworks for LLM-based GeoAI. Such frameworks must move beyond generic AI ethics to address specific challenges posed by the fusion of language models, geospatial interference, and multimodal spatial data.

Actionable governance mechanisms (from policy gaps to implementable controls). Regulatory frameworks frequently articulate high-level principles (privacy, fairness, transparency) but provide limited operational guidance for LLM-enabled GeoAI, where passive inference, scale sensitivity (MAUP), and uncertainty compounding can create high-stakes harms. To translate policy relevance into implementable mechanisms, we specify three governance building blocks that can be embedded in real GIS infrastructures and institutional workflows: (1) consent and purpose limitation tracking, (2) spatially explicit bias auditing and monitoring, and (3) spatial data validation and uncertainty reporting. Importantly, each mechanism should produce auditable artifacts (logs, reports, metadata) so that compliance is verifiable rather than aspirational.

The proposed framework (Figure 3) places "Ethical Governance for LLM-based GeoAI" at its core and organizes key principles as spokes radiating outward. These principles are: explainability, consent protocols, lifecycle transparency, adaptability and interoperability, spatial ethics training, external auditing, registries, participation, accountability, and fairness auditing. Together, they provide a structured basis for assessing and improving the governance of LLM-enabled GeoAI systems across technical, organizational, and societal contexts.

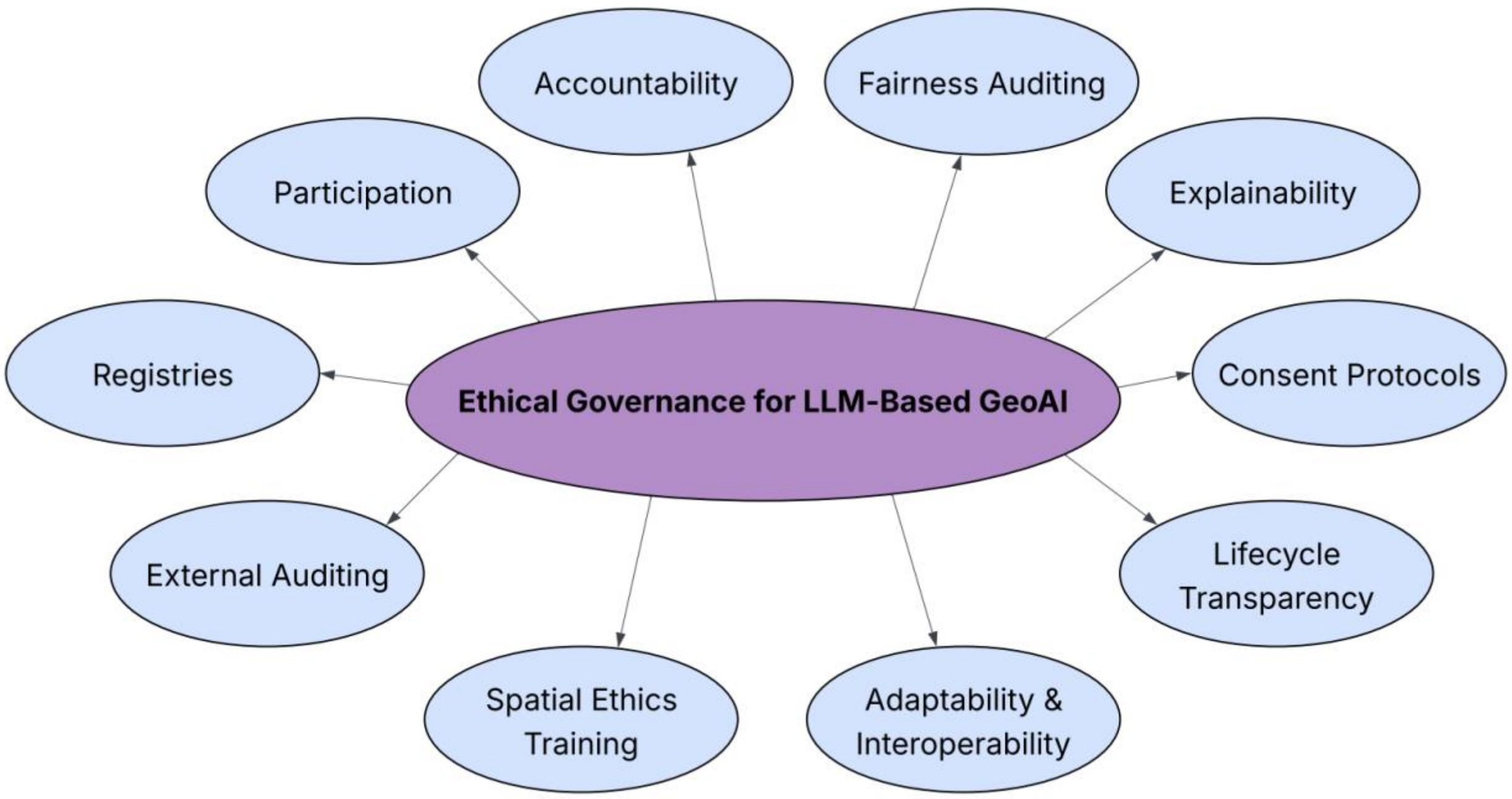

**Figure 3:** Core Principles of the Proposed Ethical Governance Framework for LLM-based GeoAI (organized as spokes around a central governance node)

To reduce ambiguity in how the governance principles in Figure 3 are enacted, we operationalize each principle as one or more enforceable system controls and institutional practices. For example, "consent protocols" are implemented as machine readable usage constraints at the data layer, with query-time enforcement and logging; "explanability" is implemented as mandatory disclosure of data recency, spatial resolution, and aggregation unit, plus tool-call traceability for LLM-mediated analyses; “external auditing” is implemented as periodic privacy and spatial fairness audits supported by retained artifacts (privacy test reports, spatially disaggregated error maps, and mitigation action logs). The operationalization enables governance to be evaluated through evidence (artifacts) rather than treated with aspirational principles.

**Table 4: Governance principles translated into enforceable controls and auditable artifacts.**

| Principle (Figure 3) | Enforceable control | Audit artifact | Validation Status |
|---|---|---|---|
| Consent protocols | Machine-readable use constraints + enforcement | Consent record + access decision logs | Candidate (proposed here; not field-tested) |
| Lifecycle transparency | Versioned lineage + prompt/tool logs | Lineage record + tool-call logs | Candidate (proposed here; not field-tested) |
| Fairness auditing | Spatially disaggregated evaluation | Spatial error maps + disparity report | Established practice (operationalizes documented GIScience diagnostic; see Table 2) |

| Principle (Figure 3) | Enforceable control | Audit artifact | Validation Status |
| --- | --- | --- | --- |
| External auditing | Scheduled privacy/fairness audits | Audit reports + mitigation records | Candidate (proposed here; not field-tested) |
| Explainability | Scale/resolution disclosure + provenance | Output metadata + provenance links | Candidate (proposed here; not field-tested) |
| Accountability | Contestation + escalation roles | Ticket logs + decision records | Candidate (proposed here; not field-tested) |

Explicitly integrated participatory processes should enable marginalized voices and local knowledge into co-designing models, monitoring their usage and development, and validation (Gupta et al., 2023). In addition, this framework should propose the creation of model and data registries for traceability, mandate external auditing for privacy risks, and require a baseline spatial training that can be further developed as a standardized process (Restall, 2024; Xing & Sieber, 2023). Finally, continued research should occur to understand how an ethical framework can be applied to different sociopolitical contexts, especially as AI becomes increasingly integrated with different parts of the world, and different levels of society. It would be helpful to work tangentially with organizations like the OECD and European Union to ensure inter-operability (*AI Principles Overview - OECD.AI*, n.d.; *Excellence and Trust in Artificial Intelligence - European Commission*,

n.d.). With these foundational principles at the core of a framework, a solid foundation can be established to build public trust for the next generation of GeoAI users.

While this paper does not present a new deployment or empirical case study, the operational controls proposed here are designed to be verifiable. In practice, evaluation would include (i) scenario based testing, (ii) spatially disaggregated performance audits to identify clustered error and disparate impact, (iii) MAUP/scale sensitivity checks for governance-relevant summaries, and (iv) privacy risk testing for inference and re-identification under realistic adversary assumptions. Each test produces auditable artifacts (logs, reports, spatial error/uncertainty surfaces) that support accountability and external review.

Stepping back, these four trade-offs share a common structure worth naming explicitly. In each case, the technically stronger option on one dimension, whether privacy (synthetic data, federated learning, minimized logging) or auditability (centralized governance, comprehensive logging, tool-first grounding), comes at a cost to a second dimension: spatial fidelity, implementation burden, or response flexibility, and none of the eight approaches discussed above has been empirically validated at scale specifically for spatial data. This pattern suggests that governance for LLM-enabled GeoAI is not simply a matter of selecting the "better" technical approach, but of making an explicit, context-dependent choice about which dimension to prioritize, privacy versus fidelity, consistency versus jurisdictional flexibility, reliability versus flexibility, and documenting that choice as part of the governance architecture itself (Section IV, Table 3), rather than treating one option as a universally correct default.

Having introduced this architecture, its enforceable controls (Tables 3 and 4), and evaluated it through the illustrative scenario above, we can now situate our contribution concretely against the frameworks discussed earlier in this section, including the operational risk-management standards named above. Table 5 summarizes this comparison.

**Table 5: Comparison of the proposed framework against principle-level, domain-literature, technical-architecture, and operational risk-management approaches.**

| Framework | Spatial specificity | LLM-specific risk coverage | Lifecycle integration | Enforceable controls | Audit artifacts | Empirical validation |
|---|---|---|---|---|---|---|
| Principle-level AI ethics frameworks (OECD AI Principles; EU Responsible AI; Floridi et al.'s AI4People) | None: spatially agnostic | None: general AI, pre-dates LLM-specific risks | No: articulates principles, not lifecycle stages | No: principles, not controls | No | N/A (normative, not empirical) |
| GeoAI ethics literature (Gupta et al., 2023; McKenzie et al., | Yes: identifies spatial risks | Partial: largely pre-LLM or general GeoAI | No: identifies risks, no system-layer mapping | No: no architectural mapping of controls | No | Varies by source; not an architecture to validate |

| Framework | Spatial specificity | LLM-specific risk coverage | Lifecycle integration | Enforceable controls | Audit artifacts | Empirical validation |
|---|---|---|---|---|---|---|
| 2023; Ntoutsi et al., 2020) | | | | | | |
| Autonomous GIS architecture papers (Li & Ning, 2023; Li et al., 2025) | Yes: technical system layers for spatial systems | No: describes capabilities, not risks/governance | Partial: system layers described, not as governance checkpoints | No: governance not embedded as a first-class layer | No | Varies; technical system papers, not risk-governance validation |
| Operational AI risk-management standards (NIST AI RMF 1.0, 2023; ISO/IEC 42001:2023) | None: general-purpose, technology/sector-agnostic | None: general AI risk categories only; no geospatial coverage | Yes: explicitly organized around the AI system lifecycle | Partial-to-Yes: NIST is voluntary/outcomes-based; ISO 42001 has certifiable Annex A controls | Yes (esp. ISO 42001) — third-party certification and audit reports | Validated only as adopted industry practice, not peer-reviewed risk-reduction efficacy |

| Framework | Spatial specificity | LLM-specific risk coverage | Lifecycle integration | Enforceable controls | Audit artifacts | Empirical validation |
|---|---|---|---|---|---|---|
| This paper's proposed architecture | High: spatial mechanisms (MAUP, autocorrelation, location inference) embedded directly | High: explicit LLM-specific risks (hallucination, prompt sensitivity, tool/agent misuse) | Yes: controls mapped to data/GIS/LLM/application layers (Figure 2) | Candidate: proposed controls specified (Tables 3–4), not yet adopted | Specified per control (Tables 3–4), not yet implemented | Not empirically validated; illustrative scenario only |

**V. Future Research Agenda for Responsible Spatial AI**

Beyond establishing an ethical governance framework proposed in Section IV, substantial research remains to be done to advance responsible LLM-based GeoAI. Given that Section III characterized the eight issues in this review as ranging from largely unaddressed to actively debated, the research priorities below are ordered roughly by urgency starting with the least mature area identified.

**LLM-specific technical risk validation (Section III.e):** Of the eight issues discussed, this is the one with the thinnest response, and it should be the first priority for future work. Spatially specific hallucination-detection methods verification approaches that check a generated spatial claim against the actual GIS layer it purports to describe do not yet exist in any form we identified in this review, despite general-purpose LLM hallucination research being well established. Similarly, prompt-robustness evaluation specific to spatial scale and aggregation-unit framing has not been studied, even though Section III.d already establishes that scale and aggregation choices materially change outputs. Future work should adapt general LLM evaluation benchmarks to include spatially explicit test cases for instance, systematically varying the areal unit specified in a prompt and measuring how much a model's narrative output changes as a result.

Future work must prioritize empirical validation of privacy preservation and bias mitigation, in diverse real-world environments (Li & Ning, 2023; Rao et al., 2023; Romano, 2025). This directly extends the gap named in Section III.c: comparative assessment of how proposed bias-auditing approaches perform when applied to genuinely spatial data where errors cluster geographically rather than distributing independently remains almost entirely absent, and understanding how these mechanisms perform across different fields is an important next step (Hochmair et al., 2025).

There is also a pressing need for the adaptation and development of explainability and interpretability tools for spatial and multimodal AI systems (Wei et al., 2025; Zhang, n.d.). Section III.f's Xing and Sieber (2023) case study demonstrated specific, documented friction when general-purpose XAI methods are applied to a spatial task; future work should treat that friction as a research agenda in its own right. Research should explore how such tools can empower those beyond technical practitioners, but also non-expert stakeholders and community members, to critically engage with outputs and identify sources of error. Developing longitudinal and participatory studies in the long run is essential to understand the downstream impacts of GeoAI decision-making, particularly in spatially disadvantaged populations (Gupta et al., 2023; Walker et al., 2023). Additionally, work is necessary to evaluate and refine educational and training programs for building AI and spatial data literacy among both professionals and the broader public (Restall, 2024; Tucker, 2024). This connects directly to the workforce-development gap identified in Section III.h: future work should design and longitudinally evaluate curricula that integrate LLM-specific risks into existing GIScience uncertainty and error-propagation training, rather than treating spatial AI literacy as a separate track from established GIS education.

Finally, as LLM-enabled Geo-AI systems are increasingly deployed in transnational contexts, comparative framework should examine interactions between technical "best practices", ethical frameworks, and heterogeneous regulatory environments found in different places globally (*AI Principles Overview - OECD.AI*, n.d.; Floridi et al., 2018). This extends the centralized-versus-decentralized governance question raised in Section III.g, special attention should be drawn to cross-border data flows, portability of accountability tools, and pathways for policy harmonization.

Addressing these research priorities will help to ensure that governance frameworks are grounded in empirical realities, adaptive to local and global contexts, and can meaningfully support the safe and equitable deployment of these systems.

## VI. Conclusions

This review has mapped eight recurring issues in the governance of LLM-enabled GeoAI i.e. spanning data provenance, spatial privacy, algorithmic bias, structural spatial mechanisms, LLM-specific technical risk, explainability, policy, and workforce development and, for each, characterized the underlying mechanism, grounded it in a documented example, and assessed how far the field's response has actually progressed. That progression varies considerably: some issues, like the spatial mechanisms discussed in Section III.d, already have established GIScience diagnostic practice waiting to be enforced as governance; others, like the LLM-specific technical risks in Section III.e, have barely been named as a distinct problem, let alone addressed.

Building on this issue-by-issue account, Section IV proposed a governance-aware architecture that maps each issue to enforceable controls and auditable artifacts across the geospatial data lifecycle. As stated there, this is not a new set of ethical principles; existing frameworks already articulate fairness, transparency, and accountability at a general level. The contribution is the explicit link between spatially specific risk mechanisms and a structure that can be audited, rather than merely aspired to.

We want to be direct about the limits of that contribution. The architecture and controls proposed here are not empirically validated; the flood-routing scenario in Section IV is illustrative, not deployed. Whether tool-first grounding actually reduces hallucination in practice, whether spatially disaggregated bias auditing catches disparities a global metric would miss, and whether the governance layer we propose survives contact with a real institutional workflow are all open questions, not settled findings. We see this as an honest reflection of where the field stands, rather than a shortcoming unique to this review: as Section III.e and III.c make clear, empirical validation is largely absent across the literature we surveyed, not only from our own proposal.

Moving forward, we believe the most productive path is the one outlined in Section V: prioritizing validation of the least mature issues first, particularly the spatially specific technical risks that general AI safety research has not yet addressed. Coordinated effort among GIS practitioners,

LLM developers, and policymakers will be needed to move governance for LLM-enabled GeoAI from principle to verified practice.